\pdfoutput=1
\documentclass[11pt]{article}

\usepackage[]{acl}

\usepackage{times}
\usepackage{latexsym}
\usepackage{graphicx}
\usepackage{algorithm}
\usepackage{algorithmic}
\usepackage{amsmath}
\usepackage{xcolor}
\usepackage{tabularx}
\usepackage{multicol}
\usepackage{booktabs}
\usepackage{hyperref}

\usepackage[T1]{fontenc}
\usepackage[utf8]{inputenc}

\usepackage{microtype}

\title{Graph Attention with Hierarchies for Multi-hop Question Answering}

\author{Yunjie He\thanks{\hspace{.5em}Work carried out as part of MSc thesis supervised by Huawei Noah's Ark Lab, London}\\
  University College London\\
  \texttt{yunjie.he.17@ucl.ac.uk} \\
  \And
  Ieva Staliūnaitė\thanks{\hspace{.5em}Work carried out while working at Huawei Noah's Ark Lab, London}\\
  Accelex Technology \\
  \texttt{ieva.staliunaite@gmail.com} \\
  \AND
  Philip John Gorinski \\
  Huawei Noah's Ark Lab, London\\
  \texttt{philip.john.gorinski@huawei.com}\\
  \And 
  Pontus Stenetorp\\
  University College London\\
  \texttt{pontus@stenetorp.se}
  }

\begin{document}
\maketitle
\begin{abstract}
Multi-hop QA (Question Answering) is the task of finding the answer to a question across multiple documents. In recent years, a number of Deep Learning-based approaches have been proposed to tackle this complex task, as well as a few standard benchmarks to assess models' Multi-hop QA capabilities. %IS I added the "QA (Question Answering)" bit so that we have defined QA before using the abbreviation
In this paper, we focus on the well-established HotpotQA benchmark dataset, which requires models to perform answer span extraction as well as support sentence prediction.
%In this paper, 
We present two extensions to the state-of-the-art Graph Neural Network (GNN) based model for HotpotQA, Hierarchical Graph Network (HGN): (i) we complete the original hierarchical structure by introducing new edges between the query and context sentence nodes; (ii) in the graph propagation step, we propose a novel extension to Hierarchical Graph Attention Network -- GATH (\textbf{G}raph \textbf{AT}tention with \textbf{H}ierarchies) -- that makes use of the graph hierarchy to update the node representations in a sequential fashion. 
%We perform experiments using the HotpotQA dataset to determine the validity of the proposed improvements. The experimental results
Experiments on HotpotQA demonstrate the efficiency of the proposed modifications and support our assumptions about the effects of model-related variables.
\end{abstract}

\section{Introduction}

Question Answering (QA) tasks can be classified into single-hop and multi-hop ones, depending on the complexity of the underlying reasoning. Different from single-hop QA %IS I would say "Contrary to" instead of "Different from"
\cite{rajpurkar-etal-2016-squad,trischler-etal-2017-newsqa,lai2017race}, where questions can be answered given a single paragraph or single sentence in the context, multi-hop QA requires us to retrieve and reason over scattered information from multiple documents, as demonstrated in Figure~\ref{fig:multihop_example}. There are many methods proposed for addressing the multi-hop QA problem. One type of these recent approaches extends well-performing single-hop machine reading comprehension models to be multi-hop, such as DecompRC \citep{min-etal-2019-multi} and QFE \citep{nishida-etal-2019-answering}.

The other avenue is to develop models specifically aimed at multi-hop QA. Among those, Graph Neural Networks (GNNs) have recently garnered a lot of attention.
%The other type of methods are to model information as a graph and use graph neural networks (GNNs) in the reasoning process.
In GNN-based approaches, gaphs are employed to represent query and context contents (nodes) and the underlying relationships between them (edges). Information between nodes is simultaneously propagated via the edges with the help of a variety of GNNs, such as Graph Convolutional Network (GCN) \citep{kipf2017semisupervised}, Graph Attention Network (GAT) \citep{velikovi2017graph}, or Graph Recurrent Network (GRN) \citep{song-etal-2018-graph}. With these GNNs, node representations are obtained conditioned on the question and context documents, and used for the QA task.

\begin{figure}[t]
    \centering
    {\small
    \begin{tabularx}{\columnwidth}{lX}
        Question: & Where did the form of music played by \textcolor{blue}{\textit{Die Rhöner Säuwäntzt}} originate? \\
        Answer: & \textcolor{red}{\underline{United States}} \\
        Supports: & \multicolumn{1}{c}{\texttt{Document 9}}\\
        &s$_1$: \textcolor{blue}{\textit{Die Rhöner Säuwäntzt}} are a \textcolor{teal}{\textbf{Skiffle}}-Bluesband from Eichenzell-Lütter in Hessen, Germany.\\
        & \multicolumn{1}{c}{\texttt{Document 4}}\\
        &s$_1$:  \textcolor{teal}{\textbf{Skiffle}} is a music genre with jazz, blues, folk and American folk influences [...]\\
        &s$_2$:  Originating as a term in the \textcolor{red}{\underline{United States}} in the first half of the 20th century [...]
    \end{tabularx}
    }
    \caption{Example of a multi-hop answer and support prediction, as found in HotpotQA.}
    \label{fig:multihop_example}
\end{figure}

In this paper, we focus on one particular GNN approach designed for the Hotpot QA benchmark, the Hierarchical Graph Network (HGN) introduced in \citet{fang-etal-2020-hierarchical}.
HGN constructs a hierarchical graph that integrates nodes from different granularity levels (question/paragraph/sentence/entity). The edges in the graph capture the interactions between the information from heterogeneous levels of the hierarchy. This hierarchical graph structure has been shown to be crucial to the model's remarkable performance\footnote{At the time of writing, HGN achieves SOTA results on HotpotQA, for GNN-based approaches.} on both finding scattered pieces of supporting information across documents and the answer span prediction. %IS I'd propose to say "dispersed"

The contribution of this work is three-fold: (i) we extend the edges of HGN with a new edge type between the query and sentences, completing its original structure; (ii) we introduce a novel extension of the Graph Attention Network -- Graph Attention with Hierarchies (GATH). GATH allows for making use of the explicit hierarchical graph structure, by propagating information through the graph in a sequential fashion based on the hierarchy's levels, rather than updating all nodes simultaneously. (iii) We perform initial experiments on the HotpotQA benchmark, providing evidence of the effectiveness of our proposed extensions.

Code related to graph completion and GATH will be made publicly available at \url{redacted}.

\section{Background}
%In this section, we provide a review of existing methods for multi-hop QA task and recently proposed Hierarchical Graph Neural Networks.

To solve the multi-hop QA problem, two general research paths have been studied. The first direction focuses on extending the successful single-hop machine reading comprehension method to the multi-hop QA. DecompRC \cite{min-etal-2019-multi} decomposes the multi-hop reasoning problem into multiple single-hop sub-questions based on span predictions and applied traditional machine reading comprehension techniques on these sub-questions to obtain answers to the question. Query-Focused Extractor (QFE) \cite{nishida-etal-2019-answering} reformulates the multi-hop QA task as a query-focused summarization task based on the extractive summarization model \cite{chen-bansal-2018-fast}.

The second research direction natively addresses the task as a multi-hop setting, and directly tries to gather the information from all context documents in order to answer the question. Many approaches based on the transformer architecture \citep{vaswani2017attention} address the multi-hop QA task as simply one of attention between all words in all available documents. In such approaches, the problem quickly becomes intractable due to the long inputs involved, and they thus typically focus on alleviating the problems of using a full attention mechanism. The Longformer \citep{beltagy2020longformer}, for example, introduces a windowed attention mechanism to localise the problem, allowing for much longer input sequences to be handled than with standard BERT-based language models \citep{devlin2018bert}.

However, recently more research effort has been put toward approaches that employ Graph Neural Networks, which allow for organising information from various sources into a graph structure before addressing the core task of Question Answering, mitigating the need for very-long-distance attention functions.

Coref-GRU \citep{dhingra-etal-2018-neural} integrates multiple evidence associated with each entity mention by incorporating co-reference information using a collection  of GRU layers of a gated-attention reader \citep{dhingra-etal-2017-gated}. However, Coref-GRU only leverages co-references local to a sentence but ignores other useful global information. To address this problem, MHQA-GRN and MHQA-GCN \cite{Song2018ExploringGP} integrate evidence in a more complex entity graph, with edges that also connect global evidence. Similarly, \citet{de-cao-etal-2019-question} also encode different relations between entity mentions in the documents and perform the graph reasoning via Graph Convolutional Network (GCN) \citep{kipf2017semisupervised}.

All of the above methods which involve Graph Neural Networks only consider entity nodes and the relations between them. The HDE-Graph \citep{tu-etal-2019-multi} extends these works by creating a new type of graph with nodes corresponding to answer candidates, documents and entities. Different edges are included into the graph to capture the interaction between these nodes. DFGN \citep{qiu-etal-2019-dynamically} constructs a dynamic entity graph and performs graph reasoning with a fusion block. This fusion block includes iterative interactions between the graph and the documents (Doc2Graph and Graph2Doc flows) in the graph construction process. Hierarchical Graph Network \cite{fang-etal-2020-hierarchical} proposes a hierarchical graph that incorporates nodes on different levels of a hierarchy, including query, paragraph, sentence, and entity nodes. This hierarchical graph allows the model to aggregate query-related data from many sources at various granularities.

%Graph neural networks \cite{10.5555/2969442.2969488, li2017gated, bianchini2001processing} are deep learning methods that work on graphs.
One limitation that all of the above conventional QA graph neural networks share is that their information propagation mechanisms do not directly utilise the (explicit or implicit) hierarchical property of the graph structure. In fields outside of Natural Language Processing, recent studies on hierarchical graph neural networks focus on passing information on each hierarchical level to the node at different attention weights. 

In multi-agent reinforcement learning, HGAT \cite{ryu2020multi} generates hierarchical state-embedding of agents. This HGAT model stacks inter-agent and inter-group graph attention networks hierarchically to capture inter-group node interaction. A two-level graph attention mechanism \cite{zhang2020relational} was developed for propagating information in the close neighborhood of each node in the constructed hierarchical graph. HATS \cite{kim2019hats} predicts stock trends using relational data on companies in the stock market. HATS selectively aggregates information from different relation types with a hierarchically designed attention mechanism. By maintaining only important information at each level, HATS efficiently filters out relations (edges) not useful for trend prediction.

However, all previous studies on hierarchical graph neural networks only exploit the possible hierarchical structure on the graph node itself. Different from the above methods, our proposed hierarchical graph attention mechanism allows the graph node embeddings to be updated in the order of the hierarchical granularity level, instead of simultaneously.

\section{Model}
As our proposed improvements are largely aimed at the established Hierarchical Graph Network (HGN) model \cite{fang-etal-2020-hierarchical} for HotpotQA, we briefly describe the original architecture. HGN builds a hierarchical graph with nodes from several granularity levels (question/paragraph/sentence/entity). This hierarchical graph structure is good at capturing the interaction between nodes from different granularity levels and has been shown beneficial to the model’s remarkable performance on both finding scattered pieces of supporting information across documents, and to answer span prediction.

%Firstly, we give a brief introduction to the important components of the HGN model.

\begin{figure*}[ht]
  \centering
  \includegraphics*[width=\textwidth]{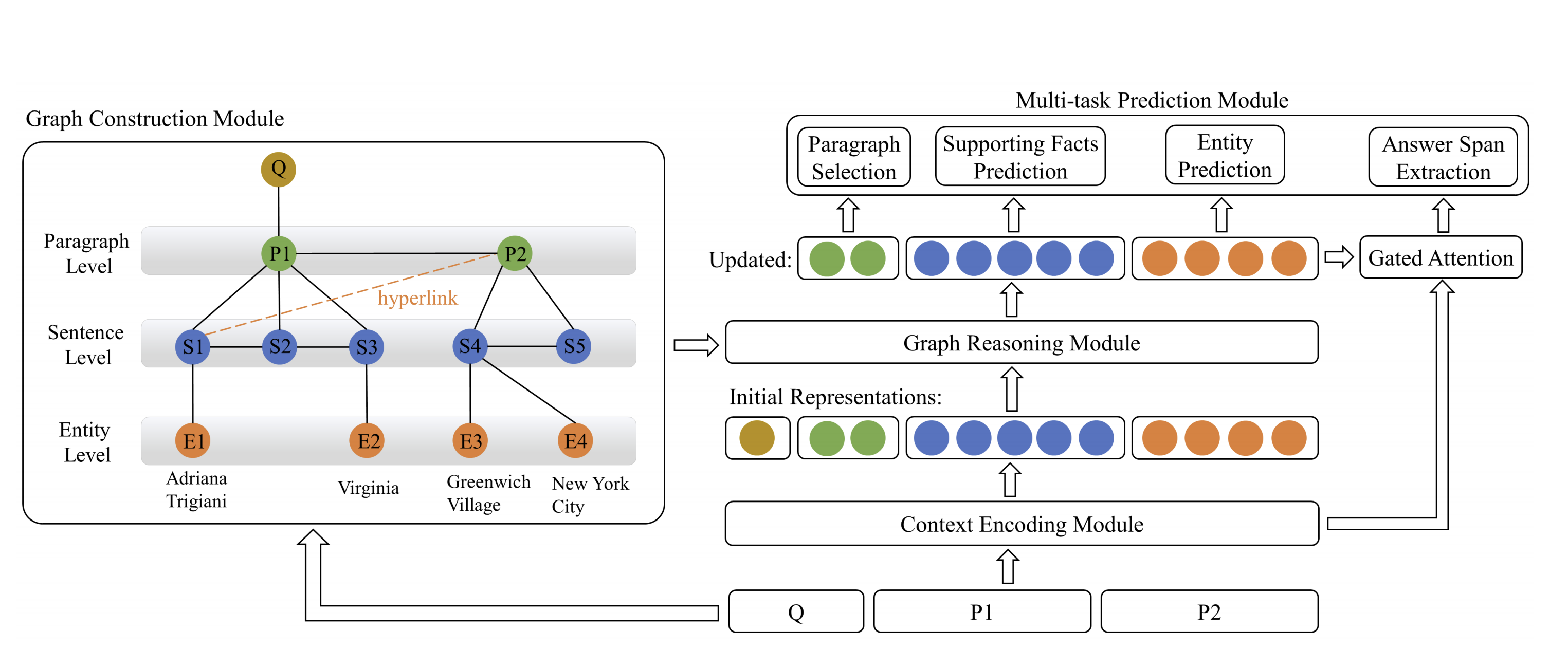}
  \caption{Model architecture of Hierarchical Graph Network (HGN). This illustration was originally introduced in \citet{fang-etal-2020-hierarchical}. We include it here for completion, to provide an overview of HGN.}
  \label{HGN architecture}
  \end{figure*}

The full HGN model pipeline consists of four modules: (i) the Graph Construction Module selects query-related paragraphs and builds a hierarchical graph that contains edges between nodes from different granularity levels within the paragraphs; (ii) the Context Encoding Module gives an initial representation/embeddings for nodes in the graph via encoding layers that consist of a RoBERTa \citep{liu2019roberta} encoder and a bi-attention layer; (iii) the Graph Reasoning Module updates the initial representation of all nodes via reasoning over the hierarchical graph; (iv) the Multi-task Prediction Module performs multiple sub-tasks including paragraph selection, supporting facts prediction, entity prediction and answer span extraction, based on the representation of all nodes. This process is summarized in Figure~\ref{HGN architecture}, as presented by the original authors of the HGN model.

We note that HGN still has limitations on its graph structure and the graph reasoning step, and in this work introduce according changes. Our proposed extensions aim to further improve HGN through a more complete graph structure, and a novel hierarchical graph nodes update mechanism. As such, our method mainly targets the Graph Construction and Graph Reasoning Modules, described in more detail below, while we leave the Context Encoding and Multi-task Prediction Modules unchanged.

\subsubsection*{Graph Construction Module}
The Hierarchical Graph is built based on the given HotpotQA question-context pair. This construction process consists of two steps: (i) multi-hop reasoning paragraph retrieval from Wikipedia, i.e. selecting candidate paragraphs with potential multi-hop relationship to the question as paragraph nodes; (ii) adding edges between question, sentence and entity nodes within the retrieved paragraphs.

In particular, the first step consists of retrieving ``first-hop'' paragraphs, that is, paragraphs of Wikipedia entries that belong to entities mentioned in the question. After this, a number of ``second-hop'' paragraphs is selected, from Wikipedia articles that are hyper-linked from these first hops.

Our work keeps the original paragraph selection method, but introduces novel meaningful edges between graph nodes.  

\subsubsection*{Context Encoding Module}
With the hierarchical graph structure in place, representations of the nodes within the graph are obtained via the Context Encoding Module. In this encoder, query and context are concatenated and fed into a pretrained RoBERTa \cite{liu2019roberta}. The obtained representations are
further passed into a bi-attention layer \cite{seo2018bidirectional} to enhance the cross interactions between the question and the context. Through this encoding mechanism, the question node is finally represented as \textbf{q} $\in$ R$^d$ and the i-th paragraph/sentence/entity nodes are represented by \textbf{p}$_i$, \textbf{s}$_i$ and \textbf{e}$_i$ $\in$ R$^d$ respectively.

\subsubsection*{Graph Reasoning Module}
Intuitively, the initial representations of the graph nodes only carry the contextualized information contained within their local contexts. To benefit from the hierarchy and information across different contexts, the Graph Reasoning Module further propagates information between the graph nodes using a single-layered Multi-head Graph Attention Network (GAT) \cite{velikovi2017graph}.
However, we believe the simultaneous node-update performed by standard GAT can be improved, in the presence of the explicitly given hierarchical property of the graph. We therefore propose a novel hierarchical graph reasoning method that performs node updates sequentially, for different levels of the hierarchy. In this manner, nodes on certain granularity levels of the graph are allowed to first aggregate some information, before passing it on to their neighbours on other levels. We speculate that this staggered information passing paradigm can be beneficial to the multi-hop Question Answering task, by passing on more question-specific contextualized information to relevant nodes.
%However, this architecture fails to capture the hierarchical property of the graph. We instead propose a novel hierarchical graph reasoning method such that the node representations can be updated in hierarchical manner. 

\subsubsection*{Multi-task Prediction Module}
The final step of the HGN model is to jointly predict answer and supporting facts for the question via multi-task learning based on the updated graph node representations. This is decomposed into five sub-tasks: (i) paragraph selection determines if a paragraph contains the ground truth; (ii) sentence selection determines if a sentence from the selected paragraph is a supporting fact; (iii) answer span prediction finds the start and end indices of the ground-truth span; (iv) answer type prediction predicts the type of the question; (v) entity prediction determines if the answer can be found among the selected entities. The above sub-tasks are jointly trained through multi-task learning with the final objective of the total loss from these sub-tasks:
\begin{align}
    \begin{split}
     \mathcal{L}_{joint} = &  \mathcal{L}_{start} + \mathcal{L}_{end} + \lambda_1\mathcal{L}_{para} +\\
     & \lambda_2\mathcal{L}_{sent} + \lambda_3\mathcal{L}_{entity} + \lambda_4\mathcal{L}_{type}
     \end{split}
\end{align}

With HGN re-introduced for completeness, we describe our proposed extensions to the original architecture in the subsequent sections.

\subsection{Completion of the graph structure}
\label{sec:completion}
HGN constructs a hierarchical graph connecting the query node with the selected multi-hop paragraphs. Each selected paragraph contains sentences and entities which are also encoded as nodes in the hierarchical graph. The graph not only incorporates the natural hierarchy existing in paragraphs, sentences and entities, but also includes helpful connections between them to faciliate the structual information propagation within the graph. Specifically, the graph consists of seven types of edges, which link the nodes in the graph. These edges are (i) edges between the question node and first-hop paragraph nodes; (ii) edges between paragraph nodes; (iii) edges between sentences in the same paragraph; (iv) edges between paragraph nodes and the corresponding within-paragraph sentence nodes; (v) edges between second-hop paragraphs and the hyperlinked sentences; (vi) edges between the question node and its matching entity nodes; (vii) edges between sentence nodes and their corresponding within-sentence entity nodes.

\begin{figure}[t]
\centering
\includegraphics*[width=\columnwidth]{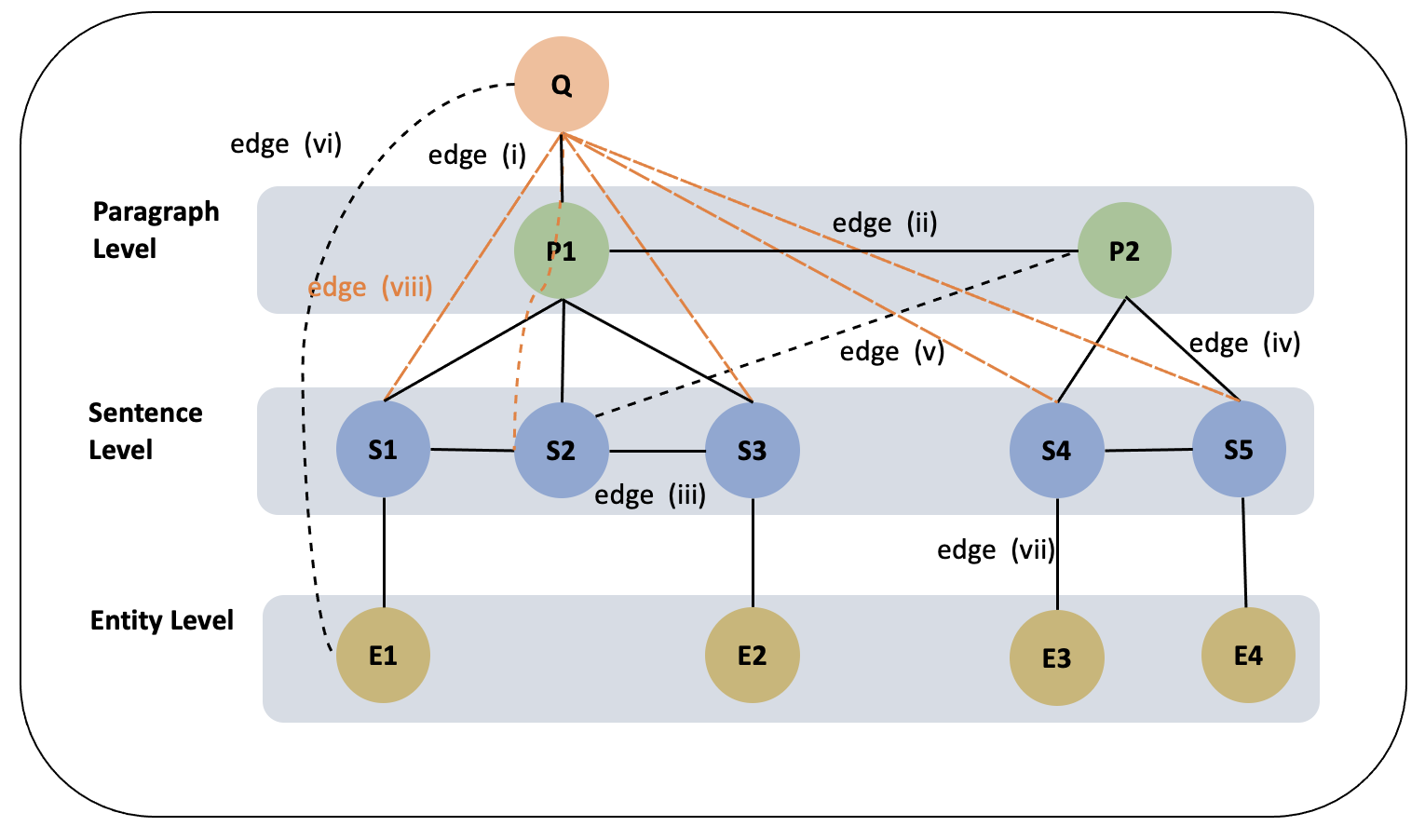}
\caption{Hierarchical Graph with (orange-colored) new question$\_$sentence edges added.}
\label{que_sent}
\end{figure}

We note that the only type of edge that seems to be missing from the graph are question-sentence edges. Hence, we first complete the hierarchical graph by introducing novel question$\_$sentence edges which connect the question node with all sentence nodes of selected paragraphs. Such new connections are introduced as edge (viii) in the hierarchical graph. The constructed hierarchical graph with novel edges added is illustrated in Figure~\ref{que_sent}. We reason that this more complete graph might help the model to learn more useful embedding because of the modification in the graph topology, which facilitates the information transmission between the question and sentences.

\subsection{Graph Attention with Hierarchies}
\label{sec:gath}
The Graph Reasoning Module updates the contextualized representations of graph nodes to capture the information aggregated from topological neighbours such that the local structures of these nodes can be included. In HGN, this process is realized by the Graph Attention Network (GAT) \citep{velikovi2017graph}, a well-established GNN approach.

However, we note that in the specific setting of Multi-hop QA with the presence of an explicit hierarchical graph structure, GAT might not be able to make full use of the information encoded in the graph, as it will not directly capture the crucial dependencies between ``levels'' of the hierarchical. To address this problem, we propose a novel Graph Attention Network with Hierarchies (GATH) which updates nodes \emph{sequentially} conditioned on an imposed order over the hierarchy levels. This is expected to help the model more effectively processes the local observation of each node into an information-condensed and contextualized state representation for individual nodes on specific levels, e.g. for paragraphs, \emph{before} passing their information on to their neighbours on other levels, such as to entity nodes. We expect this staggered flow of information might help the model aggregate information that is more useful and conditioned on the task at hand. 

The nodes in the graph are split into four categories, and can be represented by \textbf{q}, \textbf{P}, \textbf{S} and \textbf{E}: 
\[\textbf{P} = \{\textbf{p}_i\}_{i=1}^{n_p} \hspace{5mm} \textbf{S} = \{\textbf{s}_i\}_{i=1}^{n_s} \hspace{5mm} \textbf{E} = \{\textbf{e}_i\}_{i=1}^{n_e}\]
with each node embedded with an embedding function as described above, into a $d$-dimensional vector.
These node representations are jointly represent the graph nodes as
\[\textbf{H} = \{\textbf{q}, \textbf{P}, \textbf{S}, \textbf{E}\} \in R^{g\times d}, g = 1+n_p + n_s + n_e\]

GATH updates all initial node embedding \textbf{H} to \textbf{H}$^{'}$ through hierarchical graph updates. Different from GAT, GATH updates the nodes representation sequentially, according to a pre-determined order of hierarchical levels, instead of simultaneously. It takes the initial node representations \textbf{H} as input,
but first only updates information of node features of the first hierarchical level while keeping other node embeddings unchanged.

For example, if the first level to be updated is the paragraph level, we obtain the updated graph representation
\[\textbf{H}^{para} = \{\textbf{h}_1, \textbf{h}_{2}^{'}, \textbf{h}_{3}^{'}, \dots \textbf{h}_{1+n_{p}}^{'}, \textbf{h}_{2+n_{p}}, \dots , \textbf{h}_g\}\]

Specifically,
\begin{equation}
\textbf{h}_i^{'} = \mathbin\Vert_{k=1}^{K} \text{LeakyRelu(}\sum_{j\in N_i}\alpha_{ij}^{k}\textbf{h}_j \textbf{W}^{k})
\end{equation}
where $\mathbin\Vert_{k=1}^K$ represents concatenation of K heads, $\textbf{W}^k$ is the weight matrix to be learned, $N_i$ represents the set of neighbouring nodes of node i and $\alpha_{ij}^k$ is the attention coefficient calculated by:
\begin{equation}
%\resizebox{6.5cm}{!}
     {
    \alpha_{ij}^{k} = \frac{
            \text{exp(LeakyRelu}([\textbf{h}_i;\textbf{h}_j]\textbf{w}_{e_{ij}}^{k}))
        }
        {
            \sum_{t\in N_i}\text{exp(LeakyRelu(}[\textbf{h}_i;\textbf{h}_t]\textbf{w}_{e_{it}}^{k}))
        }
    }
\end{equation}
where \([\textbf{h}_i;\textbf{h}_j]\) denotes the concatenation of $\textbf{h}_i$ and $\textbf{h}_j$, and \(\textbf{w}_{e_{ij}}^{k}\) is the weight vector corresponding to the edge between node i and j.

Based on the updated embeddings on the paragraph level $\textbf{H}^{para}$, we might next consider updating the information on the sentence level. GATH propagates information to all nodes on the sentence level based on \textbf{H}$^{para}$. 
This will output a further updated graph representation
\[\textbf{H}^{sent} = \{\textbf{h}_1, \textbf{h}_{2}^{'}, \dots \textbf{h}_{1+n_{p}+n_{s}}^{'}, \textbf{h}_{2+n_{p}+n_{s}}, \dots, \textbf{h}_g\}\]
with all nodes in $\textbf{P}$ and $\textbf{S}$ updated.

\begin{figure*}[t]
\hfill
\includegraphics*[width=\textwidth]{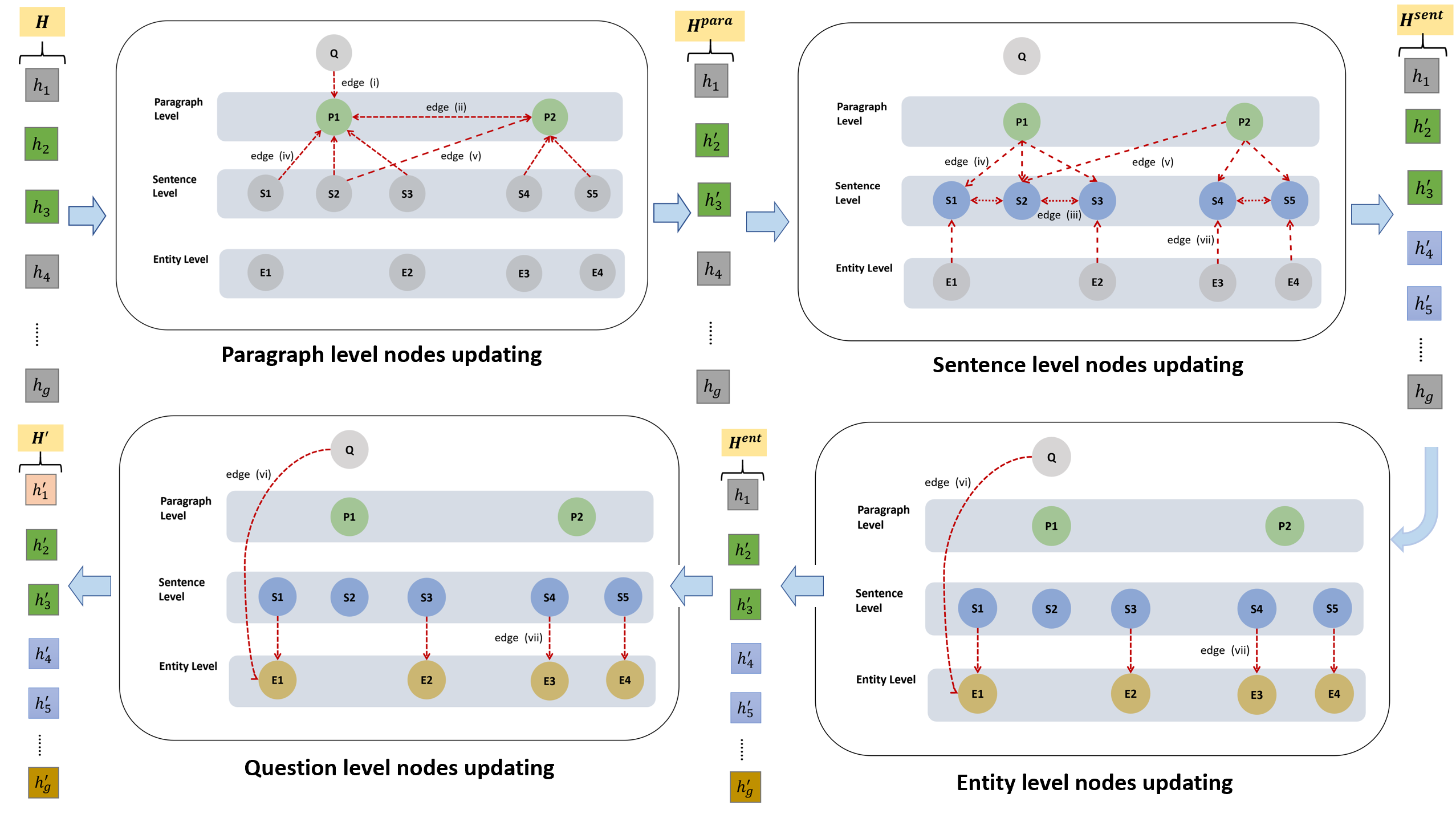}
\caption{Hierarchical node representation update process. The grey-colored graph nodes are initial contextualized embedding given by the Context Encoding Layer. Through the paragraph level message passing layer, only the neighboring information of all paragraph nodes can be passed and renewed on them. Similar steps repeat for sentence level and entity level. For convenience of labeling indices, we set $n_{p}=2$, $n_{s}=5$, $n_{e}=4$}
\label{Hierarchical node update mechanism}
\end{figure*}

Continuing the process in this manner, we eventually will have updated all node representations to obtain \textbf{H}$^{'}$ $\{\textbf{h}_1^{'},\textbf{h}_2^{'},...,\textbf{h}_g^{'}\}$. Algorithm 1 summarizes the above procedures in pseudo code. Additionally, these updating steps are combined and illustrated in Figure \ref{Hierarchical node update mechanism}.

\begin{figure}[t]
    \centering
    \includegraphics[width=7cm]{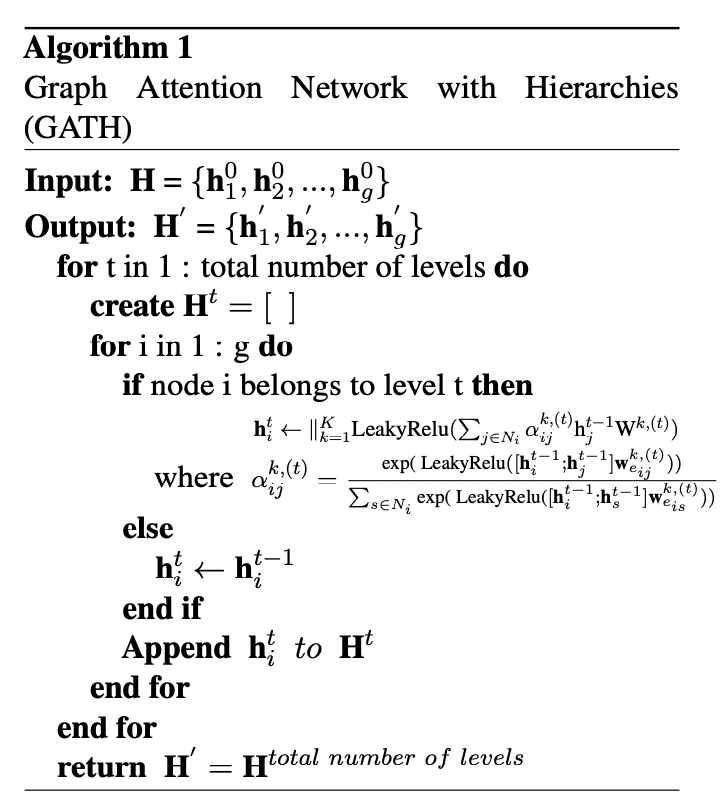}
    \label{fig:algorithm}
\end{figure}
%\begin{algorithm}[t]
%\caption{\\Graph Attention Network with Hierarchies (GATH)}
%\label{Hierarchical Graph Nodes Update Algorithm}
%\begin{algorithmic}
%\REQUIRE \textbf{H} = $\{\textbf{h}_1^{0}, \textbf{h}_2^{0}, ..., \textbf{h}_g^{0}\}$\\
%\ENSURE \textbf{H}$^{'}$ = $\{\textbf{h}_1^{'}, \textbf{h}_2^{'}, ..., \textbf{h}_g^{'}\}$
%\FOR{t in 1 : total number of levels}
%\STATE \textbf{create} $\textbf{H}^{t} = [\hspace{2mm}]$
%\FOR{i in 1 : g}
%   \IF{node i belongs to level t}
%   \hspace{5mm}
%\resizebox{5cm}{!}
%      {%
%      $\textbf{h}_{i}^{t} \gets
%\mathbin\Vert_{k=1}^{K} \text{LeakyRelu}(\sum_{j\in N_i}\alpha_{ij}^{k,(t)}\text{h}_j^{t-1} \text{W}^{k,(t)})%
%$}
%\text{where}\hspace{2mm}\resizebox{5.5cm}{!}
%      {%
%    $\alpha_{ij}^{k,(t)} 
%      = \frac{\text{exp( LeakyRelu}([\textbf{h}_i^{t-1};\textbf{h}_j^{t-1}]\textbf{w}_{e_{ij}}^{k,(t)}))}{\sum_{s\in N_i}\text{exp( LeakyRelu(}[\textbf{h}_i^{t-1};\textbf{h}_s^{t-1}]\textbf{w}_{e_{is}}^{k,(t)}))}%
%      $}
%\ELSE
%$\textbf{h}_{i}^{t} \gets \textbf{h}_{i}^{t-1}$
%\ENDIF\\
% Append
%\(\textbf{Append}\hspace{2mm}\textbf{h}_{i}^{t}\hspace{2mm}to\hspace{2mm}\textbf{H}^{t}\)
%
%\ENDFOR
%\ENDFOR\\
%\(\textbf{return}\hspace{2mm} \textbf{H}^{'} = %\textbf{H}^{total\hspace{1mm}number\hspace{1mm}of\hspace{1mm}levels}\)
%\end{algorithmic}
%\label{alg:gath}
%\end{algorithm}

\section{Experiments}
In this section, we present experiments comparing our extended HGN models with GATH with the original one employing GAT, and provide a detailed analysis of the proposed improvements and results.

For all experiments, we use RoBERTa$_{large}$ as the base embedding model. We train with a batch size of 16 and a learning rate of $1e^{-5}$ over $5$ epochs, with $\lambda_1, \lambda_3, \lambda_4=1$ and $\lambda_2=2$, and we employ a dropout rate of $0.2$ on the transformer outputs, and $0.3$ throughout the rest of the model.

\subsection{Dataset}
The effects of the above proposed improvements are assessed based on HotpotQA \cite{yang2018hotpotqa}. It is a dataset with 113k English Wikipedia-based question-answer pairs with two main features: (i) It requires reasoning over multiple documents without constraining itself to an existing knowledge base or knowledge schema; (ii) Sentence-level supporting facts are given for the answer to each question, which explain the information sources that the answer comes from. The performance of models on HotpotQA is mainly assessed on two metrics, exact match (EM) and F1 score. The model is expected to not only provide an accurate answer to the question, but also to give supporting evidences for its solution. Thus, EM and F1 score are calculated for both answer spans and supporting facts.

HotpotQA has two settings: Distractor and Fullwiki. In the distractor setting, context paragraphs consist of 2 gold truth paragraphs containing information that is needed to solve the question, and 8 paragraphs retrieved from Wikipedia based on the question, serving as related yet uninformative distractors for the question-answer pair. In the Fullwiki setting, all context paragraphs come from Wikipedia's top search results, and they need to be pre-ranked and selected in a first step. Compared with the distractor setting, this setting requires us to propose an additional paragraph selection model concerned with information retrieval, before we address multi-hop reasoning task. As all our proposed extensions aim at the graph construction and reasoning steps, we only perform these initial experiments to assess the impact of our approach in the distractor setting, where we are independent from the influence of such a retrieval system.
% the impact of the novel enhancements on the HotpotQA dataset in the distractor setting.\footnote{All changes proposed by our new method are not involved with the information retrieval system in Fullwiki setting.}

\subsection{Experimental Results}
Using the HotpotQA dataset, the models with our extensions of graph completion and GATH 
%as introduced in Sections~\ref{sec:completion} and~\ref{sec:gath}
are compared against the baseline model of HGN with standard GAT. Since it could reasonably be argued that GATH ``simulates'' a (partially) multi-layered GAT in the sense that some nodes are updated only after others have already been able to incorporate neighbouring information -- which in standard GAT requires at least two full layers -- we also include an HGN trained with a two-layer network rather than the single layer used in the original paper.

Table~\ref{Results-GraphCompletion-Full} summarizes the results on the dev set of HotpotQA\footnote{Authors' note: unfortunately, despite our best efforts we were not able to reproduce the numbers reported for HGN in \citet{fang-etal-2020-hierarchical}, even with their original, open-sourced code. We tried both the hyper-parameters as published in the paper, and the ones shipped with the code release however, the RoBERTa$_{large}$ performance when training from scratch was consistently much lower than expected on dev ($\sim74$ vs $\sim70$ joint F1). We contacted the original authors, who were not able to help out with this. In light of these discrepancies, we decided to focus only on dev set performance when assessing the impact of our extensions against re-trained vanilla HGN, as a fair comparison to the original model on test was not possible at this time.}.

\paragraph{Completion of graph structure} The HGN with new query-sentence edges improves over the baseline by 0.7/0.4 on Joint EM and F1 scores. This supports our intuition that the the missing question-to-sentence edges can indeed bring advantages to the model's abilities of both answer span extraction and supporting facts prediction.

\begin{table*}[t]
    \centering
    \begin{tabularx}{\textwidth}{lXXXX|XXXX|XXXX}
    \toprule
    & \multicolumn{4}{c}{Answer} & \multicolumn{4}{c}{Support} & \multicolumn{4}{c}{Joint}\\
    \multicolumn{1}{c}{Model} & \multicolumn{1}{c}{EM} & \multicolumn{1}{c}{F1} & \multicolumn{1}{c}{P} & \multicolumn{1}{c|}{R} & \multicolumn{1}{c}{EM} & \multicolumn{1}{c}{F1} & \multicolumn{1}{c}{P} & \multicolumn{1}{c|}{R} & \multicolumn{1}{c}{EM} & \multicolumn{1}{c}{F1} & \multicolumn{1}{c}{P} & \multicolumn{1}{c}{R} \\
    \midrule
         HGN (baseline) & 64.5 & 78.3 & 81.6 & 79.0 & 60.4 & 87.4 & 89.6 & 87.5 & 42.7 & 70.3 & 75.0 & 70.9\\
         HGN (2-layer GAT) & 64.1 & 78.0 & 81.4 & 78.9 & 59.9 & \textbf{87.7} & 89.5 & 88.4 & 41.6 & 70.0 & 74.5 & 71.3\\
         \midrule
         HGN (que\_sent edge) & 65.0 & 79.1 & 82.2 & 80.1 & 60.9 & 87.0 & 89.9 & 86.9 & 43.4 & 70.7 & 75.7 & 71.5\\
         HGN with GATH(P/S/E) & 66.1 & 80.1 & 83.1 & 81.1 & 54.2 & 80.1 & 86.1 & 79.5 & 38.7 & 66.5 & 73.6 & 66.9\\
         HGN-GATH(E/S/P) & 66.4 & 80.3 & 83.6 & 81.2 & 38.3 & 74.4 & 70.0 & \textbf{90.0} & 27.2 & 61.2 & 59.8 & 74.1\\
         HGN-GATH(S/E/P) & 67.0 & 80.6 & \textbf{83.7} & 81.4 & 60.5 & 86.3 & \textbf{92.3} & 83.7 & \textbf{43.9} & 71.5 & \textbf{78.8} & 70.2\\
         HGN-GATH(S/P/E) & \textbf{66.8} & \textbf{80.7} & 83.8 & 81.6 & 38.7 & 74.6 & 70.3 & \textbf{90.0} & 27.3 & 61.5 & 60.1 & \textbf{74.5}\\
         \midrule
         HGN (Combined) & 66.7 & \textbf{80.7} & \textbf{83.7} & \textbf{81.7} & \textbf{61.4} & 87.0 & 91.2 & 85.6 & \textbf{43.9} & \textbf{71.9} & 77.8 & 71.8\\
    \bottomrule
    \end{tabularx}
    \caption{Performance of the proposed HGN with completed edges (HGN que\_sent), GATH, and both extensions combined on the development set of HotpotQA in distractor setting, against the baseline model HGN with GAT.}
    %Using the open-sourced code provided by \cite{fang-etal-2020-hierarchical}, we are temporarily unable to obtain the same statistics reported in the paper.}
    \label{Results-GraphCompletion-Full}
\end{table*}

\paragraph{Graph Attention with Hierarchies} GATH allows for pre-defining the order of level updates in the model. Given that the order in which the hierarchy levels are updated is likely to affect the model’s performance, we perform experiments with different orders (P/S/E\footnote{P/S/E abbreviates Paragraph/Sentence/Entity}$^{,}$\footnote{We exclude the query level update to make it more comparable to the baseline model, which also excludes this update.}, E/S/P, S/E/P and S/P/E) and compare them to the baseline models with one and two-layer GAT.
All the GATH-based extended models outperform the baseline model on the answer-span extraction by an absolute gain of $1.6$ to $2.4$ points on the answer extraction metrics.

On the other hand, the order of hierarchical levels does show an influence on the model’s evidence collection ability. The ``wrong'' order leads to worse performance of the extended model, such as in the E/S/P and S/P/E cases.

On most metrics, but specifically on joint F1 score, the extended GATH-based model with the order S/E/P outperforms not only the baseline model, but also the other GATH models. It achieves a Joint EM/F1 score of $43.9/71.5$, exceeding the baseline model's performance by $1.2$ each. 

Interestingly, the 2-layer GAT version of HGN slightly under-performs when compared to the original HGN setup. While gaining $0.3$ points in support prediction F1, it loses the same amount of performance in answer prediction and joint scores. We assume this is why the original HGN calls for only one layer, when we could intuitively have expected multi-layered networks to perform better.

\paragraph{Combined query-sentence edges and GATH} The above experimental results demonstrate the individual effectiveness of these two proposed improvements of graph completion and GATH. Naturally, we are also interested in the performance resulting from combining both. The ``HGN (Combined)'' row in Table~\ref{Results-GraphCompletion-Full} represents the model combining graph completion and GATH-S/E/P. This combined model brings slight improvement over the other models on most metrics. This final model sees further improvements, particularly in the answer span prediction task, and achieves the overall highest joint F1 score at $71.9$, indicating that the contributions of graph completion and GATH are mutually benefitial.
 
\subsection{Error Analysis}
In this section, we perform an error analysis on the concrete influence of the proposed HGN (combined) model based on question types. The majority of questions in HotpotQA fall under the \emph{bridge}\footnote{requiring a bridging entity between support sentences, needed to arrive at the answer} and \emph{comparison} reasoning categories. As suggested by \citet{fang-etal-2020-hierarchical}, we split comparison questions into \emph{comp-yn} and \emph{comp-span}. The former represents questions that should answer the comparison between two entities with ``yes'' or ``no'', e.g. ``Is Obama younger than Trump?'', while the latter requires an answer span, e.g. ``Who is younger, Obama or Trump?''.

Table~\ref{Results-Reasoning} shows the performance of the original HGN model and the proposed model HGN-GATH (combined) on various types of reasoning questions. Results indicate that \emph{comp-yn} questions are easiest for both models, and the \emph{bridge} type is the hardest to solve. The analysis table shows that HGN (combined) is more effective than the original model on all of these reasoning kinds except support EM for \emph{comp-yn}, though even here the much improved answer prediction leads to an overall improvement of $2.42$ on Joint EM.

\begin{table}[ht]
\centering
{\small
\begin{tabularx}{\columnwidth}{lcccc}
\toprule
\multicolumn{5}{c}{HGN-GAT}\\
Question & Ans EM & Sup EM & Joint EM & Pct($\%$)\\
\midrule
comp-yn & 81.22 & 81.44 & 68.34 & 6.19\\
comp-span & 65.50 &  71.04 & 48.49 & 13.90\\
bridge & 63.08 & 57.01 & 39.73 & 79.91\\
\\
\multicolumn{5}{c}{HGN-GATH (Combined)}\\
    Question & Ans EM & Sup EM & Joint EM & Pct($\%$)\\
    \midrule
    comp-yn & 85.81 & 80.79 & 70.96 & 6.19\\
    comp-span & 68.42 & 71.53 & 50.34 & 13.90\\
    bridge & 64.95 & 58.08 & 40.71 & 79.91\\
    \bottomrule
    \end{tabularx}
}
\caption{Original HGN (top) and HGN-GATH combined (bottom) model results for various reasoning types. `Pct' signifies percentage of all questions per category.}
\label{Results-Reasoning}
\end{table}

\section{Conclusions and Future Work}
In this paper, we proposed two extensions to Hierarchical Graph Network (HGN) for the multi-hop Question Answering task on HotpotQA. First, we completed the hierarchical graph structure by adding new edges between the query and context sentence nodes. Second, we introduced GATH as the mechanism for neural node updates, a novel extension to GAT that can update node representations sequentially, based on hierarchical levels. To the best of our knowledge, this is the first time the hierarchical graph structure is directly exploited in the update mechanism for information propagation.

Experimental results indicate the validity of our approaches individually, as well as when used jointly for the multi-hop QA problem, outperforming the currently best performing graph neural network based model, HGN, on HotpotQA.

In the future, we would particularly like to integrate hierarchical graph attention weights into GATH, as motivated by related research in Reinforcement Learning.

\bibliographystyle{acl_natbib}
\bibliography{acl}
\end{document}